%% file: root.tex
\documentclass[letterpaper, 10 pt, conference]{ieeeconf}  

\IEEEoverridecommandlockouts                              

\usepackage{graphics} 
\usepackage{epsfig} 
\usepackage{mathptmx} 
\usepackage{times} 
\usepackage{amsmath} 
\usepackage{url}
\usepackage{amssymb}  

\title{\LARGE \bf
Real-Time Marker Localization Learning for GelStereo Tactile Sensing
}

\author{Shujuan Liu$^{1}$, Shaowei Cui$^{2}$, Chaofan Zhang$^{3}$, Yinghao Cai$^{2}$, and Shuo Wang$^{4}$
\thanks{This work was supported in part by the National Key Research and Development Program of China (No.2021ZD0114505), in part by the Beijing Natural Science Foundation under Grant 4222056, in part by the Beijing Nova Program under Grant Z211100002121152, and in part by the National Natural Science Foundation of China under U1913201. (corresponding author: Shaowei Cui, email: {\tt\small shaowei.cui@ia.ac.cn})}
\thanks{$^{1}$ Shujuan Liu is with the University of Science and Technology Beijing.}%
\thanks{$^{2}$ Shaowei Cui and Yinghao Cai are with the State Key Laboratory of Management and Control for Complex Systems, Institute of Automation, Chinese Academy of Sciences, Beijing 100190, China. }%
\thanks{$^{3}$ Chaofan Zhang is with the School of Artiﬁcial Intelligence, University 
of Chinese Academy of Sciences, Beijing 100049, China, and also with Institute of Automation, Chinese Academy of Sciences, Beijing 100190, China.}
\thanks{$^{4}$ Shuo Wang is with the State Key Laboratory of Management and Control for Complex Systems, Institute of Automation, Chinese Academy of Sciences, Beijing 100190, China, with the School of Artiﬁcial Intelligence, University of Chinese Academy of Sciences, Beijing 100049, China, and also with the Center for Excellence in Brain Science and Intelligence Technology Chinese Academy of Sciences, Shanghai 200031, China.}
}
\usepackage{graphicx}
\usepackage{epstopdf}
\usepackage{subfigure}
\usepackage{amsmath}
\usepackage{multirow}
\begin{document}

\maketitle
\thispagestyle{empty}
\pagestyle{empty}

\begin{abstract}

Visuotactile sensing technology is becoming more popular in tactile sensing, but the effectiveness of the existing marker detection/localization methods remains to be further explored. Instead of contour-based blob detection, this paper presents a learning-based marker localization network for GelStereo visuotactile sensing called Marknet.
Specifically, the Marknet presents a grid regression architecture to incorporate the distribution of the GelStereo markers. Furthermore, a marker rationality evaluator (MRE) is modelled to screen suitable prediction results. The experimental results show that the Marknet combined with MRE achieves 93.90\% precision for irregular markers in contact areas, which outperforms the traditional contour-based blob detection method by a large margin of 42.32\%. Meanwhile, the proposed learning-based marker localization method can achieve better real-time performance beyond the blob detection interface provided by the OpenCV library through GPU acceleration, which we believe will lead to considerable perceptual sensitivity gains in various robotic manipulation tasks.
\end{abstract}

\section{INTRODUCTION}

\begin{figure}[!t]
\centering  
\includegraphics[width=\linewidth]{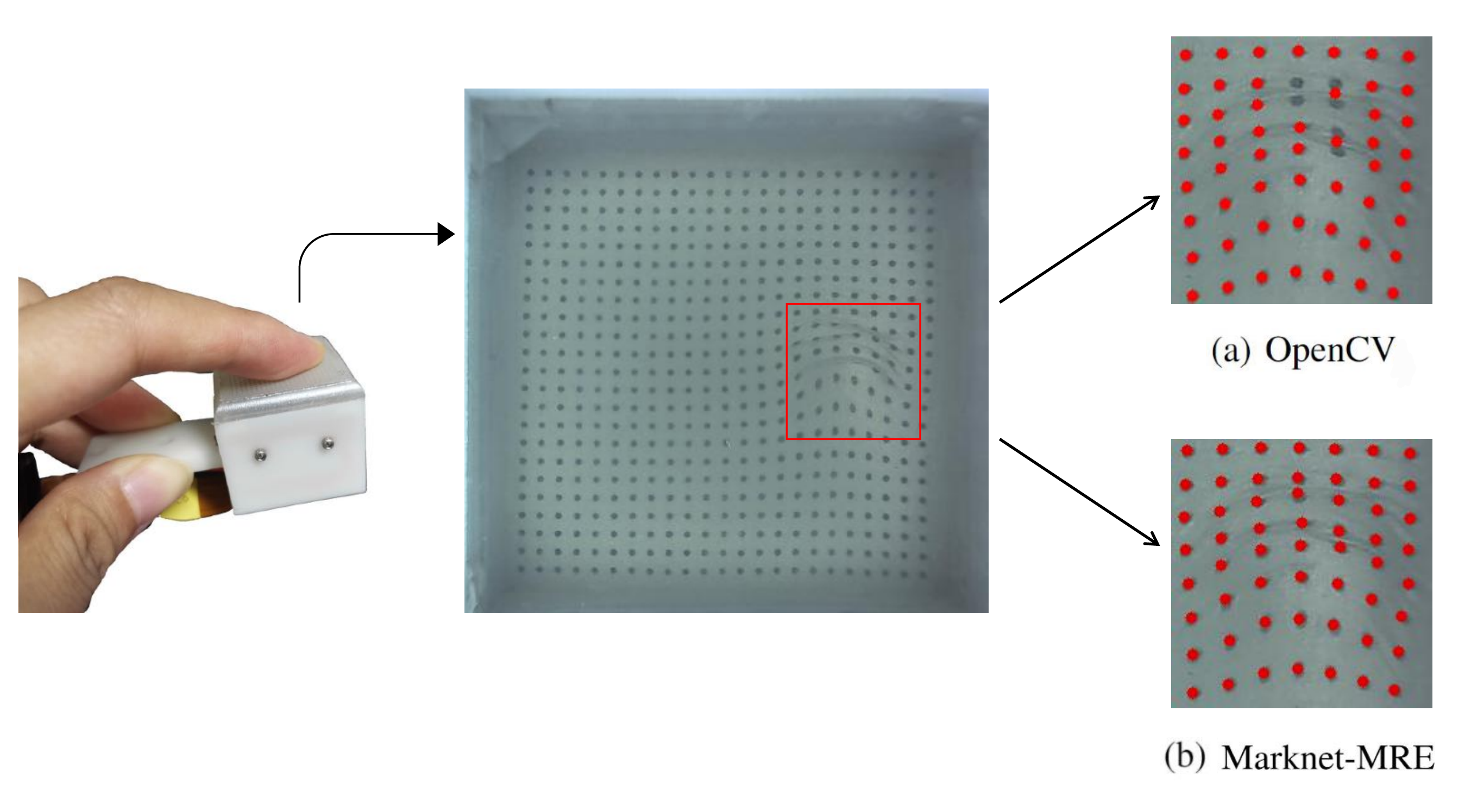} 
\caption{The marker localization results visualization of the proposed method and blob detection interface provided by OpenCV. (a) OpenCV; (b) Marknet-MRE.}  
\label{fig1}
\end{figure}

For humans and robots, the sense of touch is important for object recognition and dexterous manipulation~\cite{5339133}. When touching an object, we quickly learn a set of its physical properties, such as shape, smoothness, hardness, and thermal conductivity, which allow us to classify objects quickly and design appropriate operational strategies~\cite{johansson2009coding}. To date, researchers have developed some different tactile sensors based on various transduction principles, and a detailed review of the existing tactile sensors can be found in \cite{pyo2021recent}. Recently, camera-based tactile sensors (visuotactile sensors) have attracted much attention in robotic communities because of the benefit of high spatial resolution and multi-mode tactile sensing capabilities \cite{yuan2017gelsight,abad2020visuotactile}, such as contact geometry~\cite{cuiTIE}, slippage~\cite{dong2019maintaining}, and force~\cite{kakani2021vision}. Specifically, the visuotactile sensors capture the contact information as a tactile image and further extract tactile features based on the markers embedded in the contact gel layer. However, the existing marker detection methods mainly adopt the blob detection interface provided by OpenCV~\cite{bradski2000opencv}, while this method relies heavily on manually designed parameters and is prone to failure in the contact area, as shown in Fig. \ref{fig1}. Therefore, it is necessary to study the marker localization method with a stronger adaptive ability for visuotactile sensing.

In this paper, we try to address the marker localization problem by learning-based methods, i.e. using artificial neural networks. Naturally, there are many well-studied object detection networks~\cite{zaidi2022survey}, such as Faster R-CNN~\cite{ren2015faster}, SSD~\cite{liu2016ssd}, and YOLO~\cite{redmon2016you}, that can infer the position of the markers directly. However, these networks are specially designed for the detection task, which requires reasoning not only about the location of the target but also about the size of the target, and the size information is not necessary for the marker localization for visuotactile sensing. As a result, many well-designed network mechanisms are redundant and computationally expensive for this task, for example, the anchor mechanism~\cite{kong2020foveabox}. Meanwhile, since the markers are small and numerous, the marker localization task requires high precision and accuracy of the inference results. In contrast, the detection ability of the common detection networks for small targets is limited~\cite{yang2022querydet}.

\begin{figure*}[!t]
    \centering
    \includegraphics[width=\linewidth]{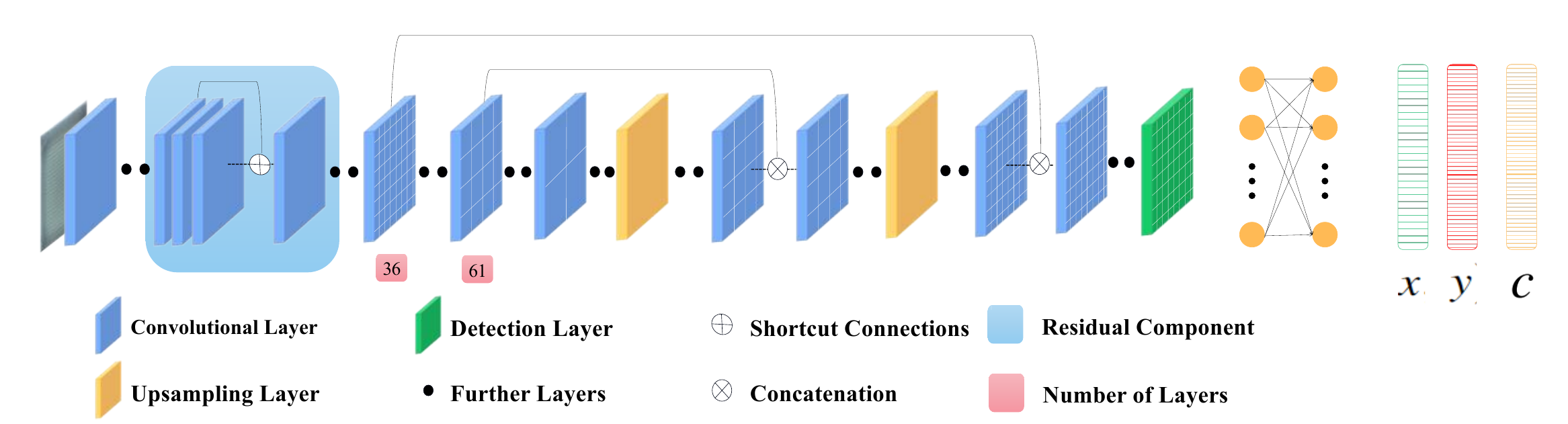}
    \caption{The network architecture of the proposed Marknet.}
    \label{fig:marknet}
\end{figure*}

To this end, we propose a learning-based marker localization network for GelStereo visuotactile sensing~\cite{cuiTIE}, which we call Marknet. Compared with traditional detection networks, the Marknet outputs the position and confidence information of the predicted markers. Inspired by the matrix-distributed markers of the GelStereo sensor, the Marknet presents a grid regression architecture to incorporate its distribution. For each grid, the network outputs several candidate points and confidence. Meanwhile, a marker rationality evaluator (MRE) is proposed to screen the final predicted marker from the above candidate results. The experimental results show that the proposed Marknet combined with MRE achieves 93.90\% precision for irregular markers in contact areas, which outperforms the traditional contour-based blob detection and YOLOv3~\cite{redmon2018yolov3} object detection methods by a
significant margin of 42.32\% and 17.94\%, respectively. An example of the markers localization results of the traditional blob detection interface provided by OpenCV and the proposed method is shown in Fig. \ref{fig1}.

\section{METHOD}
Given a tactile image $I$ captured by the GelStereo sensor, the primary goal of the marker localization method is to precisely locate all markers on the image plane. In this paper, we first generate all possible markers distributed on the tactile image, 
\begin{align}
Y_{d}=F_{d}(I) \in \mathbb{R}^{n\times 3}
\end{align}
where $F_d$ indicates the proposed Marknet to infer the potential location of the markers. Each element of $Y_d$ contains the location ($x$, $y$) and confidence ($\in (0, 1)$), and $n$ is the total number of $Y_d$.

Next, we present a Marker Rationality Evaluator (MRE) to infer the rationality of the predicted markers in $Y_d$ with low confidence.
\begin{align}
y_{e}^i=F_{e}(Y_{d}^i),\;Y_d^i\in Y_d
\end{align}
where $y_e^i$ belongs to $(0, 1)$, indicating the possibility of the predicted marker $Y_d^i$ to be a correct result.

Finally, the predicted markers are generated based on the output of the Marknet ($Y_d$) and further evaluation of MRE.

\subsection{Marknet}

The overall architecture of the Marknet model is shown in Fig. \ref{fig:marknet}, inspired by the object detection regression pipeline, i. e. YOLO~\cite{redmon2016you}, the Marknet presents a grid regression architecture to incorporate the distribution of the GelStereo markers. Specifically, the Marknet $(F_{d})$ divides the input image $(I)$ into $S\times S$ grid cells for inference. For each grid, the network outputs $m$ candidate points. Each point contains two position information $(x, y)$, which are the centre coordinates of the prediction points on the tactile image plane, and a  confidence $c$. The Marknet consists of convolutional layers, shortcut connections, upsampling layers, and detection layers. Furthermore, we observe that the GelStereo markers have similar size information. Therefore, the Marknet has only one detection layer, so the final output feature map has only one scale, which greatly simplifies computation during inference.

The loss function of Marknet consists of two parts: confidence loss and localization loss. The formula for calculating loss is as follows:
\begin{align}
L=\alpha L_{c}+\beta L_{p}
\end{align}
where $\alpha$ and $\beta$ are the weight coefficients, $L_{c}$ is the confidence loss, and $L_{p}$ is the localization loss. 
Specifically, the confidence loss is computed by binary cross-entropy loss~\cite{ruby2020binary}, and the localization loss is defined by Euclidean distance.
\begin{align}
L_{p}=\sqrt{(x_{p}-x_{t})^2+(y_{p}-y_{t})^2}
\end{align}
where $(x_{p}, y_{p})$ is the centre coordinate of the predicted point on the image plane, and $(x_{t}$, $y_{t})$ is the centre coordinate of the real marker.

When calculating the localization loss, we map $(x, y)$ to the original image size (416$\times$416) to obtain $(x_{p}, y_{p})$, and the formula is as follows:
\begin{align}
x&=\sigma (x)\times \frac{416}{S} +X_{g}\\
y&=\sigma (y)\times \frac{416}{S} +Y_{g}
\end{align}
where $(X_{g},Y_{g})$ is the coordinate of the upper left corner of the grid cell.

\subsection{Post-processing}
\begin{figure}[!t]
    \centering
    \includegraphics[width=\linewidth]{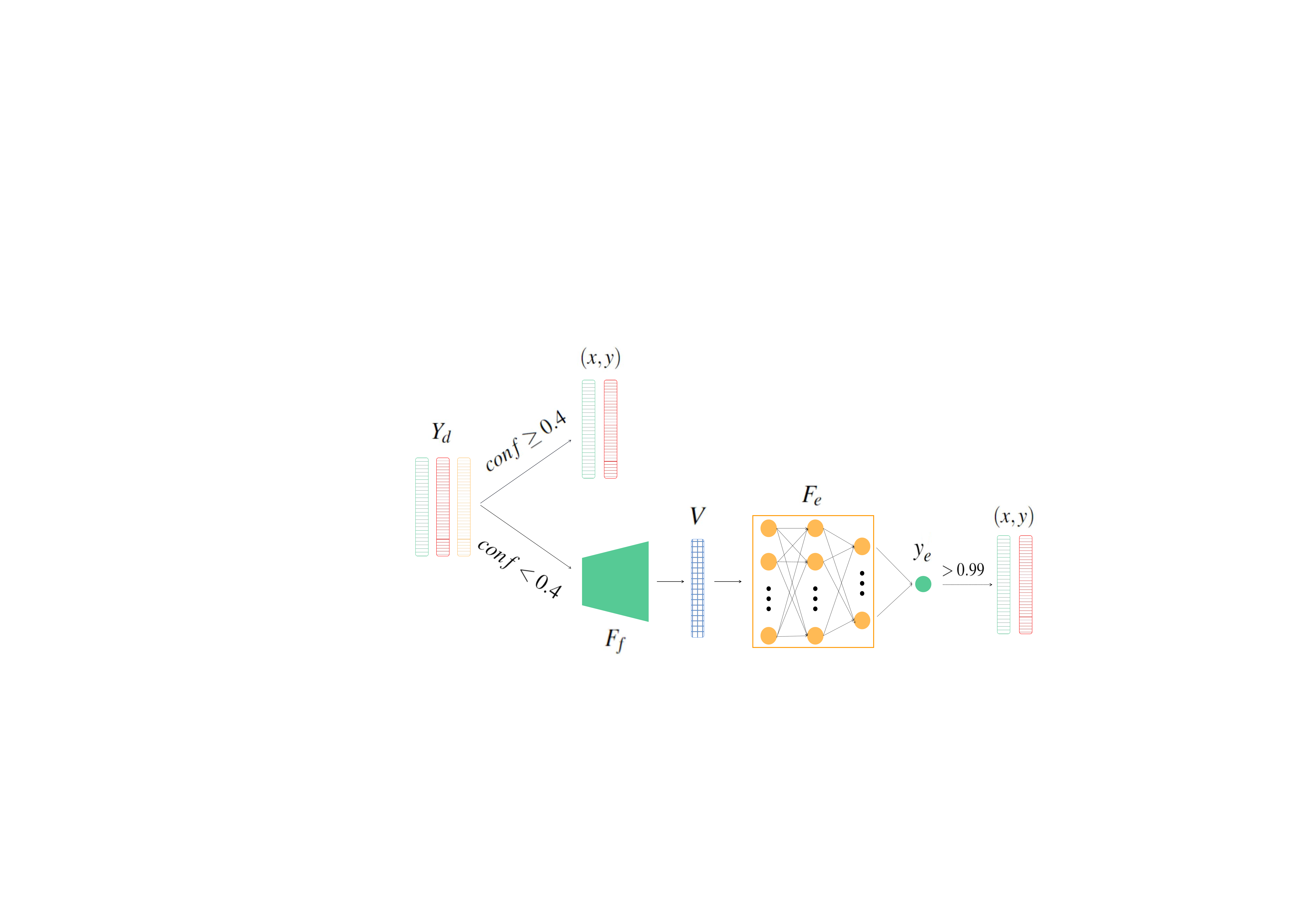}
    \caption{The post-processing pipeline of markers predicted by Marknet.}
    \label{fig:post}
\end{figure}
To screen suitable predicted markers from results $(Y_{d})$, we propose a post-processing pipeline, as shown in Fig. \ref{fig:post}. Firstly, we output the predicted markers directly when confidence is above 0.4 ($c>0.4$). Meanwhile, the MRE evaluated the remaining markers with confidence between 0.1 and 0.4 for their probability of correct estimation.

Specifically, given the predicted marker $Y_d^i$, we first construct a distance-based feature,
\begin{align}
V^i=F_{f}(Y_{d}^i)\in \mathbb{R}^{11}
\end{align}
where $V$ contains the smallest ten distances between marker $Y_d^i$ to all other predicted markers and confidence $c$. Next, feature $V^i$ is fed into the MRE ($F_{e}$), consisting of three fully connected layers, which outputs the possibility value $(y_{e}^i)$. 
\begin{align}
y_{e}^i=F_{e}(V^i)
\end{align}

The predicated marker $Y_d^i$ will be regarded as the correct marker when $y_e^i>0.99$. Note that the significance of setting the threshold of $y_e$ so high is to minimize false output, which is more critical for subsequent sensing tasks.




\section{EXPERIMENTS}

In this section, we first introduce the dataset of the marker localization network (Marknet dataset), the dataset of the MRE (MRE dataset) and the experimental setup. Next, the proposed method's comprehensive experimental evaluations are performed on these datasets.

\subsection{Dataset Introduction}
\subsubsection{Marknet dataset}
The Marknet dataset is collected by 13 objects of different sizes, shapes, and textures on the GelStereo visuotactile sensor that performs stable pressing with different strengths and sliding with different strengths and speeds. Approximately 70 to 90 images were collected per experiment. Therefore, the Marknet dataset consists of about 28,000 images, each containing 529 markers.

The method of labelling is first to use the blob detection interface provided by OpenCV\footnote{\url{https://docs.opencv.org/3.4/d0/d7a/classcv_1_1SimpleBlobDetector.html}} for rough labelling and then correct the wrong labelling or offset the labelling of the real marker through manual observation. Furthermore, the Marknet dataset is randomly divided into three parts of 6:2:2, which are used as train, validation and test datasets.

To further study the localization performance of the irregular markers in contact areas by different methods, the test dataset is divided into two parts: one named easy is labelled by OpenCV only, containing 43107 markers, and the other named hard is labelled by manual correction, containing 271 markers. Note that the markers in the hard dataset are invalid when detected by OpenCV.
\subsubsection{MRE dataset} 
The dataset for training the MRE is obtained by comparing predictions on the validation dataset with the ground truth. Specifically, the correctly predicted markers are labelled as 1, and the wrongly predicted markers are labelled as 0. For all markers predicted by the Marknet, when $c>0.4$, almost all of them are correctly predicted, which does not need the screen. So the MRE dataset only needs to contain points with $c < 0.4$. In order to make the number of correctly predicted points and correctly predicted points in the dataset equal, the latter is randomly sampled, and the final MRE dataset has 200,000 points. The ratio of the number of true positive samples and false positive samples is about 1:1.

\subsection{Experimental Setup}
\subsubsection{Baselines}
To verify the localization performance of the proposed method, we compare it to some existing methods and perform an ablation study of the proposed method:
\begin{itemize}
    \item OpenCV: The official blob detection interface provided by OpenCV~\cite{bradski2000opencv} with suitable parameters.
    \item YOLOv3: The YOLOv3 object detection network~\cite{redmon2018yolov3} with modified anchor size and output for this marker detection task.
    \item Marknet: The proposed marker localization network, and it outputs markers when confidence is above 0.2 ($c>0.2$).
    \item Marknet-MRE: The proposed marker localization network with the MRE post-processing.
\end{itemize}
\subsubsection{Network parameters comparison}
\begin{itemize}
    \item The number of grid cells ($S$): The difference in the number of grid cells on each feature map means the different receptive field sizes. The larger the number of grid cells, the smaller the receptive field and the more local and detailed the output features. The number of grid cells is selected to be $13\times13$, $26\times26$, and $52\times52$ for comparative evaluation. The training epochs are all 100 times, and each grid cell has five prediction points.
    \item The number of candidate points on each grid cell ($m$): The greater the number ($m$) of candidate points on each grid cell, the greater the probability that the marker is detected. We compare and evaluate the prediction results of 1, 2, 3, 4, and 5 on each grid cell.
\end{itemize}

\subsubsection{Performance indicator}
In order to evaluate the performance of different methods and Marknet with different parameters more comprehensively and accurately, we compute the precision, recall, and loss of the results as follows:
\begin{align}
precision&=PT/(PT+PF)\\
recall&=PT/T\\
loss&=D/PT
\end{align}
in which $(PT)$ is the number of correctly predicted points, and  $(PF)$ is the number of wrongly predicted points. $T$ represents the actual number of markers, and $D$ is the sum of the squares of the distances between all correctly predicted points and their ground-truth labels.
\begin{figure*}[!t]
 \centering

 \subfigure[Overlapping markers. Upper: OpenCV, Under: Marknet.]{
 \begin{minipage}{0.85\textwidth}
  \centering
  \includegraphics[width=1\textwidth]{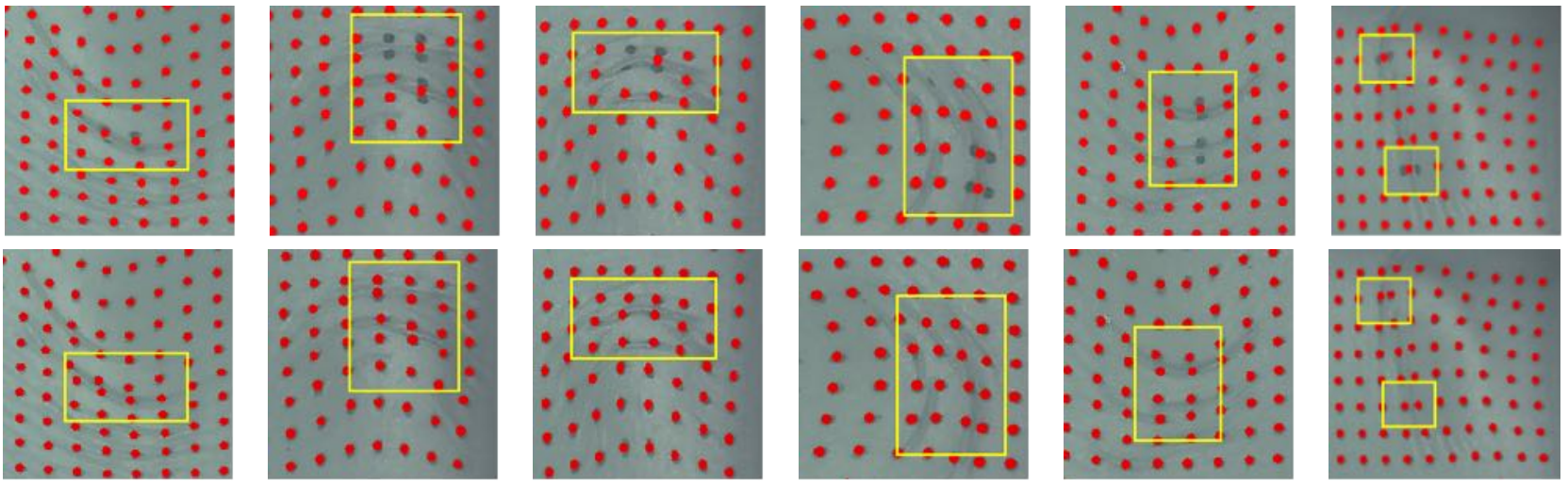}
  \label{Comparison_Overlapping}
 \end{minipage}
 }
 \subfigure[Deformed markers. Upper: OpenCV, Under: Marknet.]{

 \begin{minipage}{0.85\textwidth}
 \centering
  \includegraphics[width=1\textwidth]{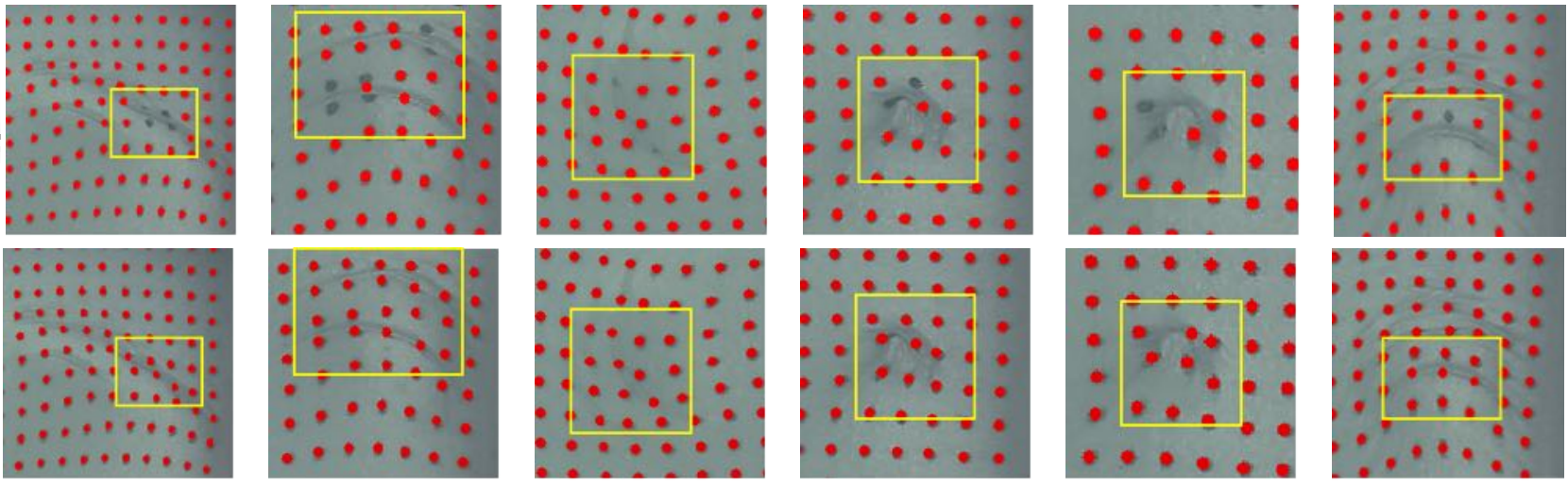}
  \label{Comparison_Deformed}
 \end{minipage}
}

 \caption{The visualization of the markers localization results provided by OpenCV and the proposed Marknet.} \label{Comparison}
\end{figure*}
\subsection{Experimental Results: different methods}

\begin{table}[!t]
    \centering
	\caption{Marker localization performance with different methods}
	\renewcommand\arraystretch{1.2}
    \begin{tabular}{c|c|ccc}
    \hline\hline
                          & method      & precision        & recall           & loss           \\ \hline\hline
    \multirow{4}{*}{hard} & OpenCV      & 51.58\%          & 18.08\%          & \textbf{1.001} \\
                          & YOLOv3      & 75.96\%          & 51.29\%          & 1.181         \\
                          & Marknet     & 93.85\% & 67.53\%          & 1.092          \\
                          & Marknet-MRE & \textbf{93.90\%}          & \textbf{73.80\%} & 1.149          \\ \hline
    \multirow{4}{*}{easy} & OpenCV      & /                & /                & /              \\
                          & YOLOv3      &95.24\%          & 95.44\%          & 0.430          \\
                          & Marknet     & \textbf{99.94\%} & 98.80\%          & \textbf{0.190} \\
                          & Marknet-MRE & 99.91\%          & \textbf{99.66\%} & 0.193          \\ \hline\hline
    \end{tabular}
    \label{tab:exp3}
\end{table}

 Table \ref{tab:exp3} summarises the localization performance of different methods on Marknet dataset. The results show that the learning-based methods outperform the traditional OpenCV method by a large margin of precision and recall on the hard dataset, which means that the learning-based methods have better generalization capability when markers are deformed. Specifically, compared with the results of OpenCV, the YOLOv3 improves the localization precision by 24.38\% and the recall by 33.21\% of deformed markers, respectively. 
 
 Compared with YOLOv3, the proposed Marknet has better localization performance, improving the localization precision by 17.89\% and recall by 16.24\%, which indicates the proposed marker localization network architecture is more suitable for our task compared to the existing object detection networks. Meanwhile, the Marknet-MRE outperforms the Marknet-only in recall performance by a margin of 6.27\%, which means that the MRE post-processing does help filter out lots of right markers.
\begin{figure}[!t]
    \centering
    \includegraphics{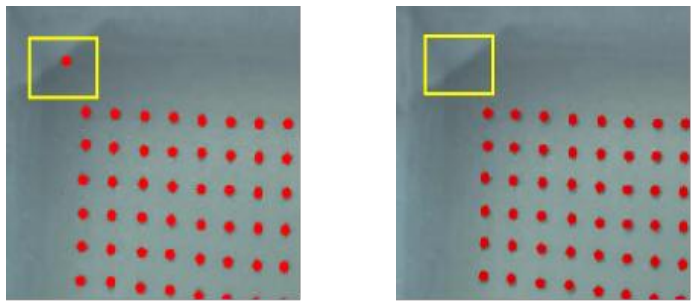}
    \caption{The outlier markers visualization. Left: OpenCV, Right: Marknet.}
    \label{Comparison_Interference}
\end{figure}

The visualization results of marker localization of the traditional blob detection interface provided by OpenCV and the Marknet-MRE are shown in Fig.\ref{Comparison}. Traditional OpenCV methods are prone to false detection, such as Fig.\ref{Comparison_Overlapping}, when two markers are close to each other and almost overlap, OpenCV will incorrectly judge them as one. Moreover, when the marker is severely deformed due to extrusion, OpenCV may have missed detection, as shown in Fig.\ref{Comparison_Deformed}. In addition, OpenCV is very susceptible to interference from surrounding noise points, and the detection range must be manually selected to filter out the interference points, as shown in Fig.\ref{Comparison_Interference}. The proposed Marknet performs better than the traditional blob detection method in the above three cases.
\subsection{Experimental Results: different network parameters}
\subsubsection{Different number of grid cells}

\begin{table}[]
    \centering
	\caption{Marker localization results with Different number of grid cells}
	\renewcommand\arraystretch{1.2}
    \begin{tabular}{c|c|ccc}
    \hline\hline
                          & $S\times S$ & precision & recall  & loss  \\ \hline\hline
    \multirow{3}{*}{hard} & 13$\times$13       & 28.57\%   & 0.74\%  & 1.715 \\
                          & 26$\times$26       & 67.37\%   & 23.62\% & 1.649 \\
                          & 52$\times$52       &\textbf{93.90\%}   &\textbf{ 73.80\%} & \textbf{1.149} \\ \hline
    \multirow{3}{*}{easy} & 13$\times$13       & 5.36\%    & 1.18\%  & 0.579 \\
                          & 26$\times$26       & 77.20\%   & 62.88\% & 0.317 \\
                          & 52$\times$52       & \textbf{99.91\%}   & \textbf{99.66\%} & \textbf{0.193} \\ \hline\hline
    \end{tabular}
    \label{tab:exp1}
\end{table}
 The experimental results are shown in Table \ref{tab:exp1}. The results show that when the number of grid cells is $52\times52$, the precision and recall rate is much higher than $13\times13$ and $26\times26$. In theory, when there is only one marker in each grid cell, the detection effect is best. This is also consistent with the experimental results. When the number of grid cells is $26 \times 26 $ or $13 \times 13 $, the step size is 16 and 32, respectively. For the marker, the grid cells are too large, and the detection effect is not satisfactory.

\subsubsection{Different number of candidate points on each grid cell}

\begin{table}[]
    \centering
	\caption{Marker localization results with different number of prediction points on each grid cell}
	\renewcommand\arraystretch{1.2}
    \begin{tabular}{c|c|ccc}
    \hline\hline
                          & $m$ & precision        & recall           & loss           \\ \hline\hline
    \multirow{5}{*}{hard} & 1   & 92.77\%          & 56.83\%          & 1.016          \\
                          & 2   & 90.10\%          & 63.84\%          &0.925          \\
                          & 3   & 92.26\%          & 61.62\%          & 0.981          \\
                          & 4   & 91.09\%          & 67.90\%          & \textbf{0.912} \\
                          & 5   & \textbf{93.90\%}          & \textbf{73.80\%} & 1.149          \\ \hline
    \multirow{5}{*}{easy} & 1   & 99.90\% & 98.88\%          & 0.167         \\
                          & 2   & 99.89\%          & 99.48\%          & 0.155        \\
                          & 3   & \textbf{99.91\%}          & 99.45\%          & \textbf{0.151} \\
                          & 4   & 99.57\%          & 99.51\%          & 0.158         \\
                          & 5   & \textbf{99.91\%}          & \textbf{99.66\%} & 0.193          \\ \hline\hline
    \end{tabular}
    \label{tab:exp2}
\end{table}
The experimental results are shown in Table \ref{tab:exp2}. The experimental results show that when there are five candidate points on each grid cell, the recall performance of the irregular marker is about 6\% to 17\% higher, and the precision is about 1\% to 4\% higher than the other four cases. Meanwhile, the detection effect of regular markers is almost the same for all five cases. The experimental results show that a relatively large number of candidate points is beneficial to improve the probability of finding potential correct markers.

\section{CONCLUSIONS}

In this paper, a learning-based marker localization network is proposed, combined with a marker plausibility evaluator to screen suitable prediction results, providing a new solution of marker localization for visuotactile sensing. Compared to the traditional blob detection method, the proposed Marknet achieves much higher localization performance (42.32\% precision, 55.72\% recall) for the irregular markers in contact areas. Meanwhile, the proposed network also outperforms the traditional object detection network by a large margin of 17.89\% precision and 16.24\% recall performance, demonstrating the effectiveness of the proposed gird-regression network architecture and the post-processing mechanism.

In the future, we will explore more efficient network architectures for marker localization to improve accuracy and real-time performance.

\addtolength{\textheight}{-12cm}   

\input{root.bbl}

\bibliographystyle{./Bib/IEEEtran}

\end{document}